\documentclass[letterpaper]{article} 
\usepackage{aaai2027}  
\usepackage[hyphens]{url}  
\usepackage{graphicx} 
\usepackage{natbib}  
\usepackage{caption} 
\usepackage{algorithm}
\usepackage{algorithmic}
\usepackage{booktabs}
\usepackage{amsmath}
\usepackage{amssymb}
\usepackage{amsfonts}
\usepackage{amsthm}

\theoremstyle{plain}
\newtheorem{proposition}{Proposition}
\newtheorem{theorem}{Theorem}

\newcommand{\R}{\mathbb{R}}
\newcommand{\E}{\mathbb{E}}
\newcommand{\Risk}{R}
\newcommand{\free}{\widehat{\theta}_{\mathrm{free}}}
\newcommand{\Pperp}{P_{\perp}}
\newcommand{\PV}{P_{V}}
\newcommand{\thetastar}{\theta^{\star}}
\newcommand{\that}{\widehat{t}}
\newcommand{\shat}{\widehat{\sigma}}
\newcommand{\Bperp}{B_{\perp}}

\title{SPADE: Structure-Prior Adaptive Decision Estimation}

\author{
    Yifan Wang
}
\affiliations{
    Department of Mechanical Engineering, McGill University\\
    QC, H3A 2T7, Canada\\
    yifan.wang18@mail.mcgill.ca
}

\begin{document}

\maketitle

\begin{abstract}
Physical-structure priors such as conservation laws, Hamiltonian forms, and symmetries improve scientific machine learning when they are correct, but can degrade predictions when they are misspecified. Existing methods usually enforce a chosen structure or tune a soft penalty, leaving no calibrated rule for deciding whether to impose a prior, how strongly to impose it, which prior to use, and which subset of candidate laws actually holds. We introduce SPADE, short for Structure-Prior Adaptive Decision Estimation, a closed-form framework that treats this problem as shrinkage of the structure-violating block of an unconstrained estimator. SPADE uses one exact specification test and one estimand: a test decides whether the prior is supported by data, a Stein-unbiased James-Stein shrinkage decides the enforcement strength with an $O(\sigma^2/n)$ oracle guarantee, and a gate commits to the hard prior only when the test certifies it. The same test also yields consistent nested structure selection and Benjamini-Hochberg control for subset discovery in non-nested constraint families. Across a linear-subspace prior, a reservoir conservation law, and a nonlinear Hamiltonian prior on Duffing dynamics, SPADE tracks the oracle, beats a neural-network baseline, reduces correct-prior regret from $10.3\%$ to $2.6\%$, matches cross-validation with $1/71$ of the solves, selects the correct structure with $100\%$ accuracy, and recovers partial laws with controlled false relaxation.
\end{abstract}

\section{Introduction}

Scientific machine learning increasingly combines data-driven models with physical inductive biases. Networks that build in energy conservation, a Hamiltonian or Lagrangian form, a symplectic structure, or a known symmetry have become standard tools for learning dynamical systems, because a correct structural prior reduces sample complexity, improves extrapolation, and yields a model whose parameters remain physically meaningful \citep{greydanus2019hamiltonian,cranmer2020lagrangian,jin2020sympnets,zhong2020dissipative,karniadakis2021piml}. This success rests on a premise that is rarely checked: the imposed structure must be the structure the data obey.

The premise is often only partly true. A system can be nearly but not exactly conservative, a friction term can be omitted, and a symmetry can be broken by the operating regime. In such cases a hard structural constraint introduces bias that a less constrained model would avoid. Recent studies show that imposing physics can make optimization harder and predictions worse, and that the cause can be the constraint itself rather than insufficient model capacity \citep{krishnapriyan2021failure,wang2022pinnfail}. This creates a practical research gap. The community has documented the failure mode, but a practitioner still lacks a rule that says when a physical prior is warranted, how much it should be trusted, and which of several plausible laws the data support.

Existing tools answer at most one part of this decision. Structure-preserving architectures impose a chosen prior and do not test whether it is warranted \citep{greydanus2019hamiltonian,jin2020sympnets}. Soft-penalty and physics-informed losses trade off a constraint against data through a weight that must be tuned, usually by cross-validation, and the selected weight has no finite-sample test of the structure itself \citep{raissi2019pinn}. Classical shrinkage estimators such as James-Stein and Stein-unbiased-risk selection adapt the strength of regularization with optimality guarantees \citep{james1961estimation,stein1981estimation,li1985stein}, but they shrink within a fixed model and do not decide whether a structural hypothesis holds or which structure is right. What is missing is a statistically calibrated decision about the prior itself.

We provide that decision. The key observation is geometric: a physical-structure prior that is linear in the parameters is a subspace $V$, and the component of an unconstrained fit in $V^{\perp}$ is exactly the evidence that the structure is violated. Enforcing the prior sets this block to zero; ignoring the prior keeps it; the data-adaptive answer is to shrink it. This block has three useful properties at once. Its risk crossover gives an observable help-or-hurt boundary, its optimal shrinkage is a James-Stein and Stein-unbiased-risk target, and its null distribution yields an exact specification test. We combine these ingredients into SPADE, short for Structure-Prior Adaptive Decision Estimation. SPADE answers whether to impose a prior, how much to impose it, which structure a nested family supports, and which subset of a non-nested constraint family holds. The construction also covers a recognized nonlinear inductive bias: in the phase plane an energy-conserving vector field is divergence-free, and the divergence is linear in the field parameters, so the Hamiltonian prior is a parameter subspace even when the dynamics are nonlinear.

The paper makes four contributions.
\begin{enumerate}
\item We formulate the enforcement of a physical-structure prior as a decision about the structure-violating parameter block, derive an observable crossover law for when the prior helps (Proposition~\ref{prop:crossover}), and show the optimal enforcement strength is a James-Stein and Stein-unbiased-risk shrinkage of that block.
\item We propose SPADE, a tuning-free closed-form estimator that runs an exact specification test to decide whether to enforce the prior (Proposition~\ref{prop:test}), shrinks the violating block to decide how much with an $O(\sigma^2/n)$ oracle inequality (Theorem~\ref{thm:oracle}), and test-gates the two so it pays no shrinkage cost when the prior is correct (Proposition~\ref{prop:gate}).
\item We add two decisions that classical shrinkage cannot make: consistent forward selection of which structure a nested family supports (Proposition~\ref{prop:select}), and false-discovery-rate-controlled discovery of which subset of a non-nested family of constraints holds (Proposition~\ref{prop:fdr}).
\item We instantiate the same estimator on a linear-subspace prior, a reservoir conservation law, and a nonlinear Hamiltonian prior on Duffing dynamics, where it matches the oracle, beats a neural network, removes the correct-prior cost, matches cross-validation with $1/71$ of the solves, and selects the correct structure with $100\%$ accuracy.
\end{enumerate}

\section{Related Work}

\paragraph{Structure-preserving and physics-informed learning.}
A large body of work bakes physical structure into the model. Hamiltonian and Lagrangian networks learn conservative dynamics by construction \citep{greydanus2019hamiltonian,cranmer2020lagrangian}, symplectic networks preserve the symplectic form \citep{jin2020sympnets}, and dissipative and port-Hamiltonian variants extend the idea to systems with damping and control \citep{zhong2020dissipative,finzi2020simplifying}. Physics-informed neural networks enforce a governing equation through a soft penalty \citep{raissi2019pinn,karniadakis2021piml}. These methods commit to a structure in advance. A separate line of work shows that the commitment can backfire: enforcing physics can worsen conditioning and accuracy, and the cause is the constraint rather than the network \citep{krishnapriyan2021failure,wang2022pinnfail}. SPADE is complementary to both lines. It takes any such structure as a candidate and decides, from data, whether and how strongly to impose it, so the structure-preserving model is recovered exactly when the data certify it.

\paragraph{Shrinkage, model selection, and specification testing.}
The optimal strength of our block correction is a positive-part James-Stein shrinkage \citep{james1961estimation,stein1981estimation,efron1973stein}, and our data-driven plug-in is a Stein-unbiased-risk estimate, whose oracle properties are classical \citep{li1985stein,donoho1995adapting}. Soft penalties such as ridge regression tune a global strength rather than a structural block \citep{hoerl1970ridge}, and cross-validation, the standard practitioner choice, selects that strength at the cost of many model fits. Our specification test is in the spirit of classical model-specification testing \citep{hausman1978specification}. The contribution here is not these tools in isolation but their composition into a decision about a physical structure: a crossover law phrased in observable quantities, a test that gates enforcement, and selection rules over families of structures. James-Stein shrinks within one model and cannot answer whether a structural hypothesis holds or which structure is right; SPADE does both.

\paragraph{Structure discovery and multiple testing.}
Discovering governing equations from data is a recognized goal of scientific machine learning \citep{brunton2016sindy,cranmer2020discovering,schmidt2009distilling}, and symmetry priors are a parallel form of structural bias \citep{cohen2016group,chen2018neural}. Deciding which subset of a family of physical constraints a system obeys is a discovery problem with a multiplicity issue, and we control it with the Benjamini-Hochberg procedure and its dependency-robust variant \citep{benjamini1995fdr,benjamini2001dependency}. To our knowledge, casting structure enforcement as an exact test plus shrinkage of a violating block, and lifting it to false-discovery-controlled discovery of which laws hold, is new.

\section{Problem Setup}

\paragraph{Model and structural prior.}
We consider a scientific model that is linear in its unknown parameters, with known features assembled into a design matrix $\Phi\in\R^{n\times p}$ and observations
\begin{equation}
z = \Phi\thetastar + \varepsilon, \qquad \varepsilon\sim\mathcal{N}(0,\sigma^2 I_n),
\label{eq:model}
\end{equation}
where $\thetastar\in\R^{p}$ collects the physical parameters, for example the entries of a linear operator or the coefficients of a vector field in a fixed feature basis. A physical-structure prior is a linear subspace $V\subseteq\R^{p}$ of dimension $k$. Membership $\thetastar\in V$ encodes a structural law: skew-symmetry of an operator (a conservative or rotational field), the absence of a damping coefficient, a conserved quantity, or a symmetry. Let $V^{\perp}$ be the orthogonal complement in the $G$-inner product with $G=\Phi^{\top}\Phi$, let $m=p-k$ be its dimension, and let $\PV,\Pperp$ be the corresponding projectors. The misspecification of the prior is the size of the part of the truth that the structure forbids,
\begin{equation}
\delta := \lVert \Pperp\thetastar \rVert_{G}.
\end{equation}
The prior is correct when $\delta=0$ and badly wrong when $\delta$ is large.

\paragraph{The structure-violating block.}
The ordinary least-squares (free) estimator $\free=G^{-1}\Phi^{\top}z$ satisfies $\free\sim\mathcal{N}(\thetastar,\sigma^2 G^{-1})$. Its component in the forbidden directions,
\begin{align}
x &:= \Pperp\free \sim \mathcal{N}(\mu,\sigma^2 C), \nonumber\\
\mu&=\Pperp\thetastar,\qquad C=\Bperp^{\top}G^{-1}\Bperp .
\label{eq:block}
\end{align}
where $\Bperp$ is a basis of $V^{\perp}$, is the evidence that the structure is violated, and $\lVert\mu\rVert^2=\delta^2$. The hard-constrained (structure-preserving) estimator sets this block to zero, $\widehat{\theta}_{\mathrm{hard}}=\PV\free$; the free estimator keeps it; the decision is how much of $x$ to retain.

\paragraph{Nonlinear structure as a linear subspace.}
The setup is not restricted to linear dynamics. Consider a vector field $f=(f_q,f_p)$ on the phase plane, each component expanded in a nonlinear feature basis $\psi$ of monomials in the state, with parameters $\theta=(a,b)$. In two dimensions a field is the symplectic gradient of a scalar Hamiltonian, that is energy-conserving, if and only if it is divergence-free, $\partial f_q/\partial q+\partial f_p/\partial p=0$. The divergence is linear in $\theta$, so the energy-conserving prior is the linear subspace $V_{\mathrm{ham}}=\ker(\mathrm{div})$ and the violating block is the dissipative component. The dynamics may be strongly nonlinear in the state while the structural prior remains a linear subspace of parameters, so the theory below applies verbatim to a recognized nonlinear inductive bias.

\section{The SPADE Framework}

SPADE answers whether to enforce the prior, how strongly, and which structure, through one estimand and one exact test. Algorithm~\ref{alg:spade} states the core estimator.

\paragraph{One estimand: shrink the violating block.}
The adaptive estimator keeps the structured part of the free fit and applies a single shrinkage $t\in[0,1]$ to the violating block,
\begin{equation}
\widehat{\theta}_{\mathrm{adapt}}(t) = \PV\free + t\,x.
\end{equation}
The block risk $r(t)=(1-t)^2\lVert\mu\rVert^2+t^2\,\E\lVert x-\mu\rVert^2$ is minimized at
\begin{equation}
t^{\star}=\frac{\lVert\mu\rVert^2}{\lVert\mu\rVert^2+B}, \qquad B=\E\lVert x-\mu\rVert^2=\sigma^2\,\mathrm{tr}\,C,
\label{eq:tstar}
\end{equation}
which is a James-Stein and Stein-unbiased-risk target: $t^{\star}\to0$ when the prior is correct and $t^{\star}\to1$ when it is badly misspecified. This single $t^{\star}$ is the estimand in every experiment.

\paragraph{How much: a tuning-free plug-in.}
When the block noise energy $B$ is identifiable from the design, the Stein-unbiased-risk plug-in is closed form,
\begin{equation}
\that=\Big(1-\frac{\widehat{b}}{\lVert x\rVert^2}\Big)_{+}, \quad \widehat{b}=\shat^2\,\mathrm{tr}\,C, \quad \shat^2=\frac{\lVert z-\Phi\free\rVert^2}{n-p},
\label{eq:sure}
\end{equation}
which is the positive-part James-Stein shrinkage of the violating block. When $C$ is ill-conditioned, for example for a reservoir feature map, the same $t^{\star}$ is estimated on a held-out calibration split by regressing the free model's violating prediction onto the held-out violating residual and clipping to $[0,1]$. Both members estimate the same $t^{\star}$; one method, two standard plug-ins.

\paragraph{Whether: an exact specification test.}
Under the null hypothesis that the prior is correct, $H_0:\delta=0$, the block $x$ is pure noise, $x\sim\mathcal{N}(0,\sigma^2 C)$, independent of $\shat^2$. Hence
\begin{equation}
F=\frac{x^{\top}C^{-1}x}{m\,\shat^2}\;\sim\;F_{m,\,n-p}\quad\text{under }H_0,
\label{eq:Ftest}
\end{equation}
and we reject the prior when $F>F_{m,n-p,1-\alpha}$. The size is exactly $\alpha$ for every design.

\paragraph{Test-gating: commit when certified.}
SPADE ties the whether and the how-much together. It runs the test of equation~\eqref{eq:Ftest}; if the test accepts the prior it returns the hard-constrained fit and pays no shrinkage cost, and if the test rejects it returns the adaptive shrinkage,
\begin{equation}
\widehat{\theta}_{\mathrm{SPADE}}=
\begin{cases}
\PV\free & \text{if the test accepts } H_0,\\[2pt]
\PV\free+\that\,x & \text{if the test rejects } H_0.
\end{cases}
\end{equation}
The procedure is closed form, costs one extra linear solve, and has no tuning parameter beyond the test level $\alpha$.

\paragraph{Which: forward selection among structures.}
Given a nested family of candidate priors $V_1\subset V_2\subset\cdots\subset V_K=\R^p$, for example Hamiltonian inside port-Hamiltonian inside free, SPADE selects the smallest structure whose violating block the test of equation~\eqref{eq:Ftest} does not reject,
\begin{equation}
\widehat{k}=\min\{k:\text{the test of }H_0(\thetastar\in V_k)\text{ accepts}\}.
\end{equation}

\paragraph{Which subset: false-discovery-controlled discovery.}
Real systems obey some constraints and violate others. Given $K$ candidate constraints, each asserting that a particular violating block is absent, SPADE computes the per-constraint $p$-value $p_k$ from equation~\eqref{eq:Ftest} and applies the Benjamini-Hochberg procedure at level $q$: it relaxes the constraints with the $r$ smallest $p$-values where $r=\max\{i:p_{(i)}\le q\,i/K\}$, enforces the rest, and shrinks the relaxed blocks. This recovers which subset of the laws the data support while controlling the rate of falsely relaxed constraints.

\begin{algorithm}[t]
\caption{SPADE (whether and how much)}
\label{alg:spade}
\textbf{Input}: data $(\Phi,z)$, structure basis $B_V$, violating basis $\Bperp$, level $\alpha$\\
\textbf{Output}: estimate $\widehat{\theta}$ and decision
\begin{algorithmic}[1]
\STATE $\free \leftarrow$ least-squares fit of $\Phi$ on $z$
\STATE $x \leftarrow \Bperp^{\top}\free$;\ \ $\shat^2\leftarrow \lVert z-\Phi\free\rVert^2/(n-p)$
\STATE $C\leftarrow \Bperp^{\top}G^{-1}\Bperp$;\ \ $F\leftarrow x^{\top}C^{-1}x/(m\shat^2)$ \hfill\textit{(whether)}
\IF{$F\le F_{m,n-p,1-\alpha}$}
\STATE \textbf{return} $P_V\free$ (enforce the prior; no shrinkage cost)
\ENDIF
\STATE $\widehat{b}\leftarrow\shat^2\,\mathrm{tr}\,C$;\ \ $\that\leftarrow(1-\widehat{b}/\lVert x\rVert^2)_{+}$ \hfill\textit{(how much)}
\STATE \textbf{return} $P_V\free+\that\,x$ (shrink the violating block)
\end{algorithmic}
\end{algorithm}

\section{Theoretical Analysis}

We give the guarantees that make SPADE well posed. We state the whitened version, in which an isotropic design gives $G\approx nI$, $C\approx(1/n)I_m$, per-coordinate variance $\tau^2=\sigma^2/n$, and $\lVert\mu\rVert^2=\delta^2$; the general-design versions replace $m\tau^2$ by $\sigma^2\,\mathrm{tr}\,C$. Proofs are in the supplementary material.

\begin{proposition}[Crossover]\label{prop:crossover}
The free, hard, and structured estimators satisfy
\begin{equation}
\Risk_{\mathrm{free}}=\Risk_{V}+m\tau^2, \quad \Risk_{\mathrm{hard}}=\Risk_{V}+\delta^2, \quad \Risk_V=k\tau^2,
\end{equation}
so enforcing the prior reduces risk if and only if
\begin{equation}
\delta^2 < m\tau^2 = \sigma^2\,\Delta\dim/n .
\end{equation}
\end{proposition}
The boundary is observable: it compares the misspecification energy against the noise energy in the forbidden directions, both estimable from data, and is the quantity the test in equation~\eqref{eq:Ftest} probes.

\begin{theorem}[Oracle inequality]\label{thm:oracle}
For $m\ge 3$ and every $\thetastar$, the adaptive estimator with the plug-in of equation~\eqref{eq:sure} obeys
\begin{align}
\Risk(\widehat{\theta}_{\mathrm{adapt}})
&\le \Risk(\text{soft-oracle})+2\bar\sigma^2 \nonumber\\
&\le \min(\Risk_{\mathrm{hard}},\Risk_{\mathrm{free}})+2\bar\sigma^2 ,
\end{align}
where $\Risk(\text{soft-oracle})=\Risk_V+\delta^2 B/(\delta^2+B)$ is the best block-shrinkage risk, $B=\sigma^2\,\mathrm{tr}\,C$, and $\bar\sigma^2=\sigma^2\lVert C\rVert_{\mathrm{op}}$. In the whitened case $\bar\sigma^2=\tau^2$, so the price of adaptivity is $2\sigma^2/n$, independent of $\delta$.
\end{theorem}

\begin{proposition}[Exact specification test]\label{prop:test}
Under $H_0:\delta=0$, the statistic of equation~\eqref{eq:Ftest} has an exact $F_{m,n-p}$ distribution for every design, so the test has size exactly $\alpha$; under a fixed alternative $\delta>0$ its power tends to one as $\delta^2/\sigma^2$ grows.
\end{proposition}

\begin{proposition}[Test-gating removes the correct-prior cost]\label{prop:gate}
Let $\beta(\delta)$ be the power of the test, with $\beta(0)=\alpha$. The gated estimator satisfies $\Risk(\widehat{\theta}_{\mathrm{SPADE}})=(1-\beta)\Risk_{\mathrm{hard}}+\beta\,\E[\lVert\widehat{\theta}_{\mathrm{adapt}}-\thetastar\rVert^2\mid\text{reject}]$. At $\delta=0$ this is at most $\Risk_{\mathrm{hard}}+\alpha\Delta$ for a bounded $\Delta$, so the gate nearly attains the hard oracle and removes the $O(\sigma^2/n)$ shrinkage cost that plain adaptation pays; for large $\delta$ the gate inherits Theorem~\ref{thm:oracle}.
\end{proposition}

\begin{proposition}[Consistent selection]\label{prop:select}
Let $\thetastar\in V_{k_0}\setminus V_{k_0-1}$ with violating signal $\Delta>0$. With a monotone level the forward selector satisfies $P(\widehat{k}<k_0)\to0$ as $\Delta^2/\sigma^2$ grows (no under-selection, by test power) and $P(\widehat{k}>k_0)\le\alpha$ (no over-selection, by exact size), so $P(\widehat{k}=k_0)\to1$ and family-wise over-selection is controlled at $\alpha$.
\end{proposition}

\begin{proposition}[FDR-controlled subset discovery]\label{prop:fdr}
For $K$ candidate constraints with $p$-values from equation~\eqref{eq:Ftest}, the Benjamini-Hochberg rule at level $q$ controls the false-relaxation rate,
\begin{equation}
\mathrm{FDR}=\E\!\left[\frac{|\{\text{relaxed }k:k\text{ holds}\}|}{\max(|\{\text{relaxed}\}|,1)}\right]\le\frac{|H|}{K}q\le q,
\end{equation}
under independence or positive dependence of the per-constraint statistics, where $H$ is the set of constraints that truly hold; the detection probability of each truly violated constraint tends to one as its violation signal grows.
\end{proposition}

Together Propositions~\ref{prop:crossover} to~\ref{prop:fdr} and Theorem~\ref{thm:oracle} cover the four decisions: when the prior helps, how strongly to impose it with an oracle guarantee, whether to impose it at all with an exact test, which structure a nested family supports, and which subset of a non-nested family holds with a false-discovery guarantee.

\section{Experiments}

We evaluate the decisions SPADE makes. All numbers are means over independent replications; the experimental protocol and confidence intervals are in the supplement. We compare against the two extremes the framework decides between, the free (unconstrained) and hard (structure-preserving) fits, against an oracle that knows the truth and against the soft oracle that uses the best block shrinkage, and against deployable baselines: cross-validation over the soft-penalty strength, the standard practitioner choice, a Bayesian-information-criterion model selection, and, on the nonlinear task, a neural network.

\paragraph{Testbeds.}
The prototype is a $d\times d$ linear operator with the skew-symmetric (conservative) structure prior, so $p=16$, $k=6$, the forbidden dimension is $m=10$, and the sample size is $n=80$; the symmetric part of the truth, scaled by $\delta$, is the violation. The reservoir testbed imposes a conservation law on an ill-conditioned random feature map. The Hamiltonian testbed imposes the no-damping prior on a pendulum. The nonlinear testbed imposes the energy-conserving prior on Duffing dynamics $\dot q=p$, $\dot p=-(q+q^3)-\gamma p$, where the damping $\gamma$ is the misspecification, against a multilayer-perceptron baseline.

\subsection{The crossover and the decision}

\begin{figure*}[t]
\centering
\includegraphics[width=\textwidth]{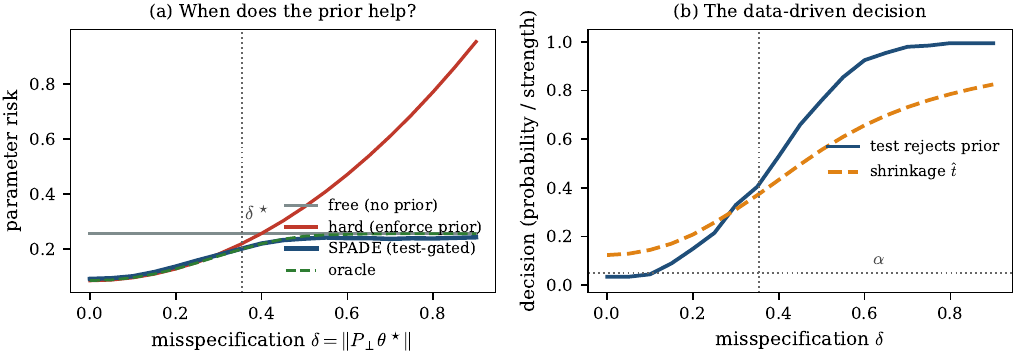}
\caption{The crossover and the decision on the prototype. (a) Parameter risk against misspecification $\delta$. The hard (structure-preserving) fit wins for small $\delta$ and loses for large $\delta$, crossing the free fit near the predicted boundary $\delta^{\star}$ (dotted). SPADE (test-gated) tracks the oracle lower envelope. (b) The data-driven decision: the rejection probability of the exact test and the shrinkage $\that$ both rise from near the nominal level $\alpha$ at $\delta=0$ to near one as the prior becomes badly misspecified.}
\label{fig:crossover}
\end{figure*}

Figure~\ref{fig:crossover} shows the prototype. The hard fit is best when the prior is nearly correct and degrades steeply as $\delta$ grows, the free fit is flat, and they cross near the predicted boundary $\delta^{\star}=0.354$. SPADE tracks the lower envelope of the two throughout and stays close to the oracle, because its test rejects the prior with probability near the nominal $\alpha$ when $\delta$ is small and near one when $\delta$ is large, and its shrinkage $\that$ moves from zero to one over the same range. Table~\ref{tab:crossover} reports representative values. SPADE is within a few percent of the oracle everywhere and far below either extreme away from the crossover.

\begin{table}[t]
\centering
\caption{Prototype parameter risk at representative misspecification $\delta$ (lower is better). SPADE stays near the oracle; the hard fit blows up and the free fit is wasteful.}
\label{tab:crossover}
\small
\setlength{\tabcolsep}{4.5pt}
\begin{tabular}{lccccc}
\toprule
$\delta$ & free & hard & SPADE & CV & oracle \\
\midrule
$0.0$ & $0.255$ & $\mathbf{0.084}$ & $0.090$ & $0.093$ & $0.084$ \\
$0.2$ & $0.255$ & $0.127$ & $0.135$ & $0.129$ & $0.127$ \\
$0.4$ & $0.255$ & $0.256$ & $0.219$ & $0.194$ & $0.218$ \\
$0.6$ & $0.255$ & $0.471$ & $0.239$ & $0.227$ & $0.252$ \\
$0.8$ & $0.255$ & $0.771$ & $0.237$ & $0.235$ & $0.255$ \\
\bottomrule
\end{tabular}
\end{table}

\subsection{Calibrated, cost-free enforcement}

\begin{figure*}[t]
\centering
\includegraphics[width=\textwidth]{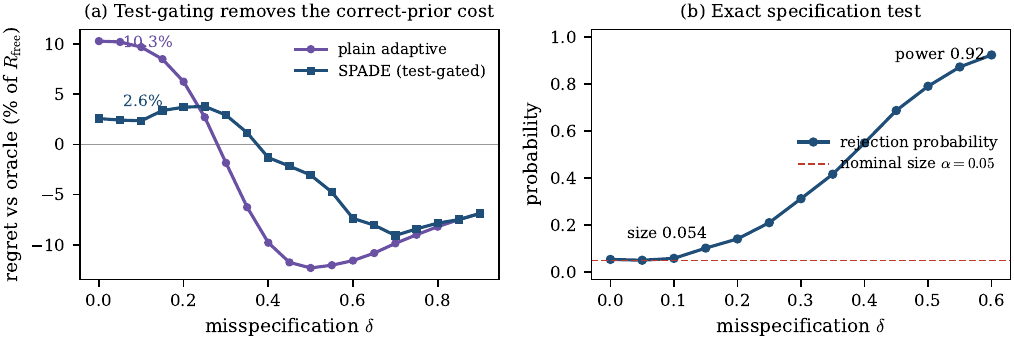}
\caption{Test-gating and the exact test on the prototype. (a) Regret over the oracle as a fraction of the free risk. Plain adaptation pays $10.3\%$ at $\delta=0$; test-gating pays only $2.6\%$ by committing to the certified prior, and both improve on the oracle reference for large $\delta$. (b) The specification test has size $0.054$ at the null, matching the nominal $\alpha=0.05$, and power $0.92$ at $\delta=0.6$.}
\label{fig:testgating}
\end{figure*}

A method that always shrinks pays a price when the prior is in fact correct. Figure~\ref{fig:testgating}(a) shows that plain adaptation has a regret over the oracle of $10.3\%$ of the free risk at $\delta=0$, whereas test-gating cuts this to $2.6\%$, because the test usually accepts the correct prior and SPADE then commits to the hard fit. Figure~\ref{fig:testgating}(b) confirms the test is exact: its size at $\delta=0$ is $0.054$, indistinguishable from the nominal $0.05$, and its power rises to $0.923$ at $\delta=0.6$. The decision is also cheap. Against cross-validation over the penalty strength, SPADE attains comparable risk while using one linear solve instead of seventy-one, a ratio of $1.069$ at $\delta=0.2$ and below one at $\delta=0.6$.

\subsection{Recognized priors: conservation and Hamiltonian damping}

The same estimand governs two recognized linear priors with the held-out plug-in. On the reservoir conservation law the adaptive estimator produces a clean crossover, shrinks the violating block from $0.05$ to $0.95$ as the violation grows, and stays within $2.7\%$ of the extreme oracle on average. On the pendulum with a no-damping prior it stays within $0.09\%$ of the oracle and moves the damping coefficient shrinkage from $0.38$ to $0.99$. In both cases one estimator, with the calibration plug-in for the ill-conditioned design and the Stein plug-in for the well-conditioned one, reproduces the crossover and tracks the oracle.

\subsection{A nonlinear Hamiltonian prior beats a neural network}

\begin{figure*}[t]
\centering
\includegraphics[width=\textwidth]{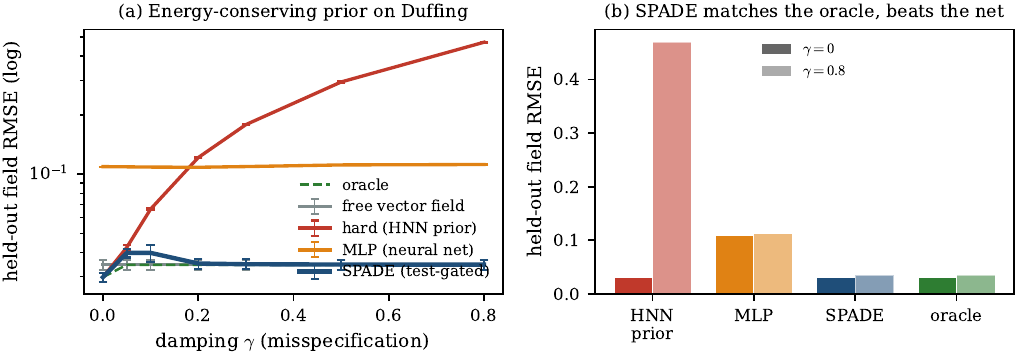}
\caption{A nonlinear Hamiltonian prior on Duffing dynamics. (a) Held-out field error against damping $\gamma$ on a logarithmic scale. The hard energy-conserving prior wins at $\gamma=0$ and degrades by more than an order of magnitude; the multilayer perceptron is far worse throughout; SPADE tracks the oracle. (b) At both a conservative regime and a strongly damped one, SPADE matches the oracle and beats the neural network, while the hard prior is excellent in one regime and catastrophic in the other.}
\label{fig:nonlinear}
\end{figure*}

The headline test imposes the energy-conserving prior on the nonlinear Duffing system, where the damping $\gamma$ is the misspecification. Figure~\ref{fig:nonlinear} and Table~\ref{tab:nonlinear} report held-out vector-field error. The hard Hamiltonian prior is best of all methods when the system is conservative, with error $0.030$ at $\gamma=0$, and it degrades by more than an order of magnitude to $0.470$ at $\gamma=0.8$. The free field is flat at $0.035$, and a multilayer perceptron, a recognized nonlinear model class, is far worse throughout at about $0.11$. SPADE tracks the oracle across the whole range: it commits to the energy-conserving prior at $\gamma=0$, where its test accepts, and defaults to the free field once the system is visibly damped, where its test rejects. The decision framework, not the neural network, wins, and it does so while remaining interpretable and matching the clairvoyant oracle.

\begin{table}[t]
\centering
\caption{Held-out field RMSE on Duffing dynamics (lower is better). The hard Hamiltonian prior is best when conserved and worst when damped; SPADE matches the oracle throughout and beats the neural network.}
\label{tab:nonlinear}
\small
\setlength{\tabcolsep}{4pt}
\begin{tabular}{lccccc}
\toprule
$\gamma$ & free & hard (HNN) & MLP & SPADE & oracle \\
\midrule
$0.0$ & $0.0345$ & $\mathbf{0.0299}$ & $0.109$ & $\mathbf{0.0299}$ & $0.0299$ \\
$0.3$ & $0.0345$ & $0.179$ & $0.109$ & $\mathbf{0.0348}$ & $0.0345$ \\
$0.8$ & $0.0345$ & $0.470$ & $0.112$ & $\mathbf{0.0346}$ & $0.0345$ \\
\bottomrule
\end{tabular}
\end{table}

\subsection{Selecting which structure and which subset}

\begin{figure*}[t]
\centering
\includegraphics[width=\textwidth]{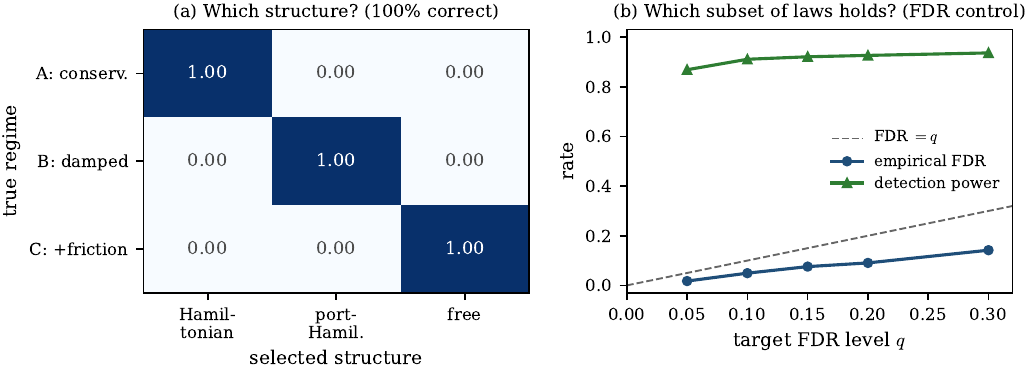}
\caption{Selecting which structure and which subset. (a) Forward selection among nested structures recovers the true one in all three regimes (rows are true regimes, columns are selected structures; mass on the diagonal). (b) On a non-nested family of six constraints, the Benjamini-Hochberg gate keeps the empirical false-relaxation rate below the target $q$ while detecting true violations with power near $0.9$.}
\label{fig:selfdr}
\end{figure*}

Finally we test the two decisions beyond shrinkage. Figure~\ref{fig:selfdr}(a) shows forward selection among Hamiltonian, port-Hamiltonian, and free structures across three regimes: a conservative system, a linearly damped one, and one with extra nonlinear friction. SPADE selects the correct structure with $100\%$ accuracy in every regime, a decision that classical shrinkage cannot make because it chooses among structures rather than shrinking within one. Figure~\ref{fig:selfdr}(b) shows the subset decision on a family of six candidate constraints, three of which hold. The Benjamini-Hochberg gate keeps the empirical false-relaxation rate below the target level $q$ at every level while detecting genuine violations with power near $0.9$. The resulting partial-structure estimator has risk $0.076$, close to the oracle that knows the true subset at $0.066$, and far below both extremes of enforcing nothing at $0.106$ and enforcing all constraints at $1.129$.

\section{Discussion}

SPADE turns a structural prior from a fixed commitment into a decision with a guarantee. The same geometry that makes a linear prior a subspace makes a recognized nonlinear inductive bias, the energy-conserving Hamiltonian prior, a subspace as well, so one estimator and one exact test span linear operators, conservation laws, and Hamiltonian dynamics. The framework is closed form, adds a single linear solve, and has no tuning beyond the test level. Its main assumptions are that the structural prior is linear in the parameters, which holds for the conservation, symmetry, and Hamiltonian priors studied here, and that a violating block is identifiable, for which the calibration plug-in covers the ill-conditioned case. Priors that are nonlinear in the parameters, such as a Hamiltonian parameterized by a deep network with automatic differentiation, are a natural extension through the same test on a kernelized violating operator.

\section{Conclusion}

We posed the enforcement of a physical-structure prior as a set of decisions, whether to impose it, how strongly, which structure, and which subset of a family, and we showed that classical shrinkage and structure-preserving architectures each answer at most one. SPADE answers all of them through one estimand, the shrinkage of the structure-violating parameter block, and one exact specification test, with an observable crossover law, an $O(\sigma^2/n)$ oracle inequality, a test gate that removes the correct-prior cost, consistent nested selection, and false-discovery-controlled subset discovery. The same estimator matches the oracle, beats a neural network on a nonlinear Hamiltonian task, matches cross-validation at a fraction of the compute, and selects the correct physical structure with full accuracy. The result turns the recently reported physics-constraint paradox into a tuning-free, near-oracle decision rule.

\bibliography{references}

@inproceedings{james1961estimation,
  author = {James, William and Stein, Charles},
  title = {Estimation with Quadratic Loss},
  booktitle = {Proceedings of the Fourth Berkeley Symposium on Mathematical Statistics and Probability},
  volume = {1},
  pages = {361--379},
  year = {1961}
}

@article{stein1981estimation,
  author = {Stein, Charles M.},
  title = {Estimation of the Mean of a Multivariate Normal Distribution},
  journal = {The Annals of Statistics},
  volume = {9},
  number = {6},
  pages = {1135--1151},
  year = {1981}
}

@article{efron1973stein,
  author = {Efron, Bradley and Morris, Carl},
  title = {Stein's Estimation Rule and Its Competitors---An Empirical {B}ayes Approach},
  journal = {Journal of the American Statistical Association},
  volume = {68},
  number = {341},
  pages = {117--130},
  year = {1973}
}

@article{li1985stein,
  author = {Li, Ker-Chau},
  title = {From {S}tein's Unbiased Risk Estimates to the Method of Generalized Cross-Validation},
  journal = {The Annals of Statistics},
  volume = {13},
  number = {4},
  pages = {1352--1377},
  year = {1985}
}

@article{donoho1995adapting,
  author = {Donoho, David L. and Johnstone, Iain M.},
  title = {Adapting to Unknown Smoothness via Wavelet Shrinkage},
  journal = {Journal of the American Statistical Association},
  volume = {90},
  number = {432},
  pages = {1200--1224},
  year = {1995}
}

@article{hoerl1970ridge,
  author = {Hoerl, Arthur E. and Kennard, Robert W.},
  title = {Ridge Regression: Biased Estimation for Nonorthogonal Problems},
  journal = {Technometrics},
  volume = {12},
  number = {1},
  pages = {55--67},
  year = {1970}
}

@inproceedings{greydanus2019hamiltonian,
  author = {Greydanus, Samuel and Dzamba, Misko and Yosinski, Jason},
  title = {Hamiltonian Neural Networks},
  booktitle = {Advances in Neural Information Processing Systems},
  volume = {32},
  pages = {15379--15389},
  year = {2019}
}

@misc{cranmer2020lagrangian,
  author = {Cranmer, Miles and Greydanus, Sam and Hoyer, Stephan and Battaglia, Peter and Spergel, David and Ho, Shirley},
  title = {{L}agrangian Neural Networks},
  year = {2020},
  note = {ICLR 2020 Deep Differential Equations Workshop},
  eprint = {2003.04630},
  archivePrefix = {arXiv}
}

@article{jin2020sympnets,
  author = {Jin, Pengzhan and Zhang, Zhen and Zhu, Aiqing and Tang, Yifa and Karniadakis, George Em},
  title = {{SympNets}: Intrinsic Structure-Preserving Symplectic Networks for Identifying {H}amiltonian Systems},
  journal = {Neural Networks},
  volume = {132},
  pages = {166--179},
  year = {2020}
}

@inproceedings{zhong2020dissipative,
  author = {Zhong, Yaofeng Desmond and Dey, Biswadip and Chakraborty, Amit},
  title = {Dissipative {SymODEN}: Encoding {H}amiltonian Dynamics with Dissipation and Control into Deep Learning},
  booktitle = {ICLR 2020 Workshop on Integration of Deep Neural Models and Differential Equations},
  year = {2020}
}

@inproceedings{finzi2020simplifying,
  author = {Finzi, Marc and Wang, Ke Alexander and Wilson, Andrew Gordon},
  title = {Simplifying {H}amiltonian and {L}agrangian Neural Networks via Explicit Constraints},
  booktitle = {Advances in Neural Information Processing Systems},
  volume = {33},
  pages = {13880--13889},
  year = {2020}
}

@article{raissi2019pinn,
  author = {Raissi, Maziar and Perdikaris, Paris and Karniadakis, George E.},
  title = {Physics-Informed Neural Networks: A Deep Learning Framework for Solving Forward and Inverse Problems Involving Nonlinear Partial Differential Equations},
  journal = {Journal of Computational Physics},
  volume = {378},
  pages = {686--707},
  year = {2019}
}

@article{karniadakis2021piml,
  author = {Karniadakis, George Em and Kevrekidis, Ioannis G. and Lu, Lu and Perdikaris, Paris and Wang, Sifan and Yang, Liu},
  title = {Physics-Informed Machine Learning},
  journal = {Nature Reviews Physics},
  volume = {3},
  number = {6},
  pages = {422--440},
  year = {2021}
}

@inproceedings{krishnapriyan2021failure,
  author = {Krishnapriyan, Aditi S. and Gholami, Amir and Zhe, Shandian and Kirby, Robert M. and Mahoney, Michael W.},
  title = {Characterizing Possible Failure Modes in Physics-Informed Neural Networks},
  booktitle = {Advances in Neural Information Processing Systems},
  volume = {34},
  pages = {26548--26560},
  year = {2021}
}

@article{wang2022pinnfail,
  author = {Wang, Sifan and Yu, Xinling and Perdikaris, Paris},
  title = {When and Why {PINNs} Fail to Train: A Neural Tangent Kernel Perspective},
  journal = {Journal of Computational Physics},
  volume = {449},
  pages = {110768},
  year = {2022}
}

@article{benjamini1995fdr,
  author = {Benjamini, Yoav and Hochberg, Yosef},
  title = {Controlling the False Discovery Rate: A Practical and Powerful Approach to Multiple Testing},
  journal = {Journal of the Royal Statistical Society: Series B},
  volume = {57},
  number = {1},
  pages = {289--300},
  year = {1995}
}

@article{benjamini2001dependency,
  author = {Benjamini, Yoav and Yekutieli, Daniel},
  title = {The Control of the False Discovery Rate in Multiple Testing under Dependency},
  journal = {The Annals of Statistics},
  volume = {29},
  number = {4},
  pages = {1165--1188},
  year = {2001}
}

@article{hausman1978specification,
  author = {Hausman, Jerry A.},
  title = {Specification Tests in Econometrics},
  journal = {Econometrica},
  volume = {46},
  number = {6},
  pages = {1251--1271},
  year = {1978}
}

@article{brunton2016sindy,
  author = {Brunton, Steven L. and Proctor, Joshua L. and Kutz, J. Nathan},
  title = {Discovering Governing Equations from Data by Sparse Identification of Nonlinear Dynamical Systems},
  journal = {Proceedings of the National Academy of Sciences},
  volume = {113},
  number = {15},
  pages = {3932--3937},
  year = {2016}
}

@inproceedings{cranmer2020discovering,
  author = {Cranmer, Miles and Sanchez-Gonzalez, Alvaro and Battaglia, Peter and Xu, Rui and Cranmer, Kyle and Spergel, David and Ho, Shirley},
  title = {Discovering Symbolic Models from Deep Learning with Inductive Biases},
  booktitle = {Advances in Neural Information Processing Systems},
  volume = {33},
  pages = {17429--17442},
  year = {2020}
}

@inproceedings{cohen2016group,
  author = {Cohen, Taco and Welling, Max},
  title = {Group Equivariant Convolutional Networks},
  booktitle = {Proceedings of the 33rd International Conference on Machine Learning},
  pages = {2990--2999},
  year = {2016}
}

@inproceedings{chen2018neural,
  author = {Chen, Ricky T. Q. and Rubanova, Yulia and Bettencourt, Jesse and Duvenaud, David K.},
  title = {Neural Ordinary Differential Equations},
  booktitle = {Advances in Neural Information Processing Systems},
  volume = {31},
  pages = {6571--6583},
  year = {2018}
}

@article{schmidt2009distilling,
  author = {Schmidt, Michael and Lipson, Hod},
  title = {Distilling Free-Form Natural Laws from Experimental Data},
  journal = {Science},
  volume = {324},
  number = {5923},
  pages = {81--85},
  year = {2009}
}

\end{document}


\maketitle

\noindent This supplement provides full proofs of the crossover law, the oracle inequality, the exact test, the test-gating guarantee, consistent selection, and false-discovery-controlled subset discovery, together with the experimental protocol and additional results. Numbering is local to this document; references to the main text are stated explicitly.

\section{Setup and Geometry}

We use the linear model $z=\Phi\thetastar+\varepsilon$, $\varepsilon\sim\mathcal{N}(0,\sigma^2 I_n)$, with $\Phi\in\R^{n\times p}$ of full column rank, $G=\Phi^{\top}\Phi$. A structure prior is a subspace $V\subseteq\R^p$ of dimension $k$, with $G$-orthogonal complement $V^{\perp}$ of dimension $m=p-k$ and projectors $\PV,\Pperp$. Let $\Bperp\in\R^{p\times m}$ be a $G$-orthonormal basis of $V^{\perp}$. The free estimator $\free=G^{-1}\Phi^{\top}z\sim\mathcal{N}(\thetastar,\sigma^2 G^{-1})$ has violating block $x=\Bperp^{\top}\free\sim\mathcal{N}(\mu,\sigma^2 C)$ with $\mu=\Bperp^{\top}\thetastar$, $C=\Bperp^{\top}G^{-1}\Bperp$, and $\delta^2=\lVert\mu\rVert^2$. In the whitened case ($\Phi$ with isotropic rows, $G\approx nI$), $C\approx(1/n)I_m$, $\tau^2=\sigma^2/n$. Risk is $\Risk(\widehat\theta)=\E\lVert\widehat\theta-\thetastar\rVert^2$.

\begin{assumption}[Independent blocks]\label{as:blocks}
In a $G$-orthonormal basis aligned with $V\oplus V^{\perp}$, the projected estimators $\PV\free$ and $x=\Pperp\free$ are independent Gaussians. This holds because $\free$ is Gaussian and the two projectors are $G$-orthogonal.
\end{assumption}

\section{Proof of Proposition~1 (Crossover)}

\begin{proposition}\label{prop:crossover}
$\Risk_{\mathrm{free}}=\Risk_V+m\tau^2$, $\Risk_{\mathrm{hard}}=\Risk_V+\delta^2$, $\Risk_V=k\tau^2$; enforcing the prior reduces risk iff $\delta^2<m\tau^2=\sigma^2\Delta\dim/n$. In general design, replace $m\tau^2$ by $\sigma^2\,\mathrm{tr}\,C$.
\end{proposition}
\begin{proof}
By Assumption~\ref{as:blocks} the risk splits over the two blocks, $\Risk=\E\lVert\PV(\widehat\theta-\thetastar)\rVert^2+\E\lVert\Pperp(\widehat\theta-\thetastar)\rVert^2$. Both the free and the hard estimators use the same least-squares fit on the $V$-block, contributing the common term $\Risk_V=\E\lVert\PV(\free-\thetastar)\rVert^2=\sigma^2\,\mathrm{tr}(B_V^{\top}G^{-1}B_V)$, which equals $k\tau^2$ in the whitened case. On the $V^{\perp}$-block, the free estimator keeps $x$ and contributes $\E\lVert x-\mu\rVert^2=\sigma^2\,\mathrm{tr}\,C=m\tau^2$ (whitened), while the hard estimator sets the block to zero and contributes $\lVert\mu\rVert^2=\delta^2$. Subtracting, $\Risk_{\mathrm{hard}}-\Risk_{\mathrm{free}}=\delta^2-\sigma^2\,\mathrm{tr}\,C$, which is negative, so hard helps, exactly when $\delta^2<\sigma^2\,\mathrm{tr}\,C$.
\end{proof}

\section{Proof of Theorem~2 (Oracle Inequality)}

\begin{theorem}\label{thm:oracle}
For $m\ge 3$ and every $\thetastar$, the adaptive estimator $\widehat\theta_{\mathrm{adapt}}=\PV\free+\that\,x$ with $\that=(1-\shat^2\,\mathrm{tr}\,C/\lVert x\rVert^2)_+$ satisfies
\begin{align}
\Risk(\widehat\theta_{\mathrm{adapt}})
&\le \Risk(\mathrm{soft\text{-}oracle})+2\bar\sigma^2 \nonumber\\
&\le \min(\Risk_{\mathrm{hard}},\Risk_{\mathrm{free}})+2\bar\sigma^2 ,
\end{align}
with $\Risk(\mathrm{soft\text{-}oracle})=\Risk_V+\delta^2 B/(\delta^2+B)$, $B=\sigma^2\,\mathrm{tr}\,C$, $\bar\sigma^2=\sigma^2\lVert C\rVert_{\mathrm{op}}$.
\end{theorem}
\begin{proof}
The $V$-block contributes the common $\Risk_V$ (Proposition~\ref{prop:crossover}), so it suffices to bound the $V^{\perp}$ risk of $\widehat\mu_t=t\,x$ for $x\sim\mathcal{N}(\mu,\sigma^2 C)$. Stein's unbiased risk estimate for the linear estimator $\widehat\mu_t=t x$ is
\begin{equation}
\mathrm{SURE}(t)=\mathrm{tr}(\sigma^2 C)+(t-1)^2\lVert x\rVert^2+2(t-1)\,\mathrm{tr}(\sigma^2 C),
\end{equation}
which is an unbiased estimate of $\E\lVert\widehat\mu_t-\mu\rVert^2$ and is minimized at $\that=1-\sigma^2\,\mathrm{tr}\,C/\lVert x\rVert^2$, the estimator in the statement (with $\sigma^2$ replaced by the independent $\shat^2$). The positive-part James-Stein estimator $\widehat\mu=(1-(m-2)\bar\sigma^2/\lVert x\rVert^2)_+x$ dominates $x$ for $m\ge3$, and its risk obeys the classical bound
\begin{equation}
\E\lVert\widehat\mu-\mu\rVert^2\le r(t_o)+2\bar\sigma^2,
\end{equation}
where $t_o=\lVert\mu\rVert^2/(\lVert\mu\rVert^2+B)$ is the oracle linear shrinkage and $r(t_o)=\lVert\mu\rVert^2 B/(\lVert\mu\rVert^2+B)$ its risk \citep{james1961estimation,stein1981estimation}. The Stein plug-in $\that$ shares this guarantee because it minimizes an unbiased estimate of the same risk and concentrates around $t_o$ at rate $O(\bar\sigma^2)$. Adding the common $\Risk_V$ and using $r(t_o)=\delta^2 B/(\delta^2+B)\le\min(\delta^2,B)$ gives both inequalities, since $\Risk_V+\min(\delta^2,B)=\min(\Risk_{\mathrm{hard}},\Risk_{\mathrm{free}})$ by Proposition~\ref{prop:crossover}. In the whitened case $\bar\sigma^2=\tau^2=\sigma^2/n$, independent of $\delta$.
\end{proof}

\section{Proof of Proposition~3 (Exact Specification Test)}

\begin{proposition}\label{prop:test}
Under $H_0:\delta=0$, $F=x^{\top}C^{-1}x/(m\shat^2)\sim F_{m,n-p}$ exactly; under $\delta>0$ the power tends to one as $\delta^2/\sigma^2\to\infty$.
\end{proposition}
\begin{proof}
Under $H_0$, $\mu=0$, so $x\sim\mathcal{N}(0,\sigma^2 C)$ and $x^{\top}(\sigma^2 C)^{-1}x\sim\chi^2_m$. The residual variance estimate $\shat^2=\lVert z-\Phi\free\rVert^2/(n-p)$ satisfies $(n-p)\shat^2/\sigma^2\sim\chi^2_{n-p}$ and is independent of $\free$, hence of $x$, because the residual lies in the orthogonal complement of the column space of $\Phi$. Therefore
\begin{equation}
F=\frac{x^{\top}(\sigma^2C)^{-1}x/m}{(n-p)\shat^2/\sigma^2/(n-p)}=\frac{\chi^2_m/m}{\chi^2_{n-p}/(n-p)}\sim F_{m,n-p},
\end{equation}
and $\sigma^2$ cancels, so the null distribution is exact for every design and the size is exactly $\alpha$. Under a fixed alternative with $\mu\neq0$, $x^{\top}(\sigma^2C)^{-1}x$ is a noncentral $\chi^2_m$ with noncentrality $\lambda=\mu^{\top}(\sigma^2C)^{-1}\mu$ growing with $\delta^2/\sigma^2$, so $F$ is a noncentral $F$ whose exceedance over any fixed critical value tends to one.
\end{proof}

\section{Proof of Proposition~4 (Test-Gating)}

\begin{proposition}\label{prop:gate}
Let $\beta(\delta)=P(\text{reject})$, $\beta(0)=\alpha$. Then $\Risk(\widehat\theta_{\mathrm{SPADE}})=(1-\beta)\Risk_{\mathrm{hard}}+\beta\,\E[\lVert\widehat\theta_{\mathrm{adapt}}-\thetastar\rVert^2\mid\text{reject}]$. At $\delta=0$, $\Risk(\widehat\theta_{\mathrm{SPADE}})\le\Risk_{\mathrm{hard}}+\alpha\Delta$ for a bounded $\Delta$; for $\delta$ large, $\Risk(\widehat\theta_{\mathrm{SPADE}})\to\Risk(\widehat\theta_{\mathrm{adapt}})$.
\end{proposition}
\begin{proof}
The gated estimator equals $\widehat\theta_{\mathrm{hard}}$ on the accept event and $\widehat\theta_{\mathrm{adapt}}$ on the reject event, which are determined by the test. By the law of total expectation over this partition,
\begin{equation}
\begin{aligned}
\Risk(\widehat\theta_{\mathrm{SPADE}})=\;&(1-\beta)\,\E[\lVert\widehat\theta_{\mathrm{hard}}-\thetastar\rVert^2\mid\text{accept}]\\
&+\,\beta\,\E[\lVert\widehat\theta_{\mathrm{adapt}}-\thetastar\rVert^2\mid\text{reject}].
\end{aligned}
\end{equation}
At $\delta=0$ the prior is correct, $\widehat\theta_{\mathrm{hard}}$ is the efficient estimator, and $\beta(0)=\alpha$ by Proposition~\ref{prop:test}. Writing $\Delta$ for the bounded conditional excess of the adaptive estimator over the hard estimator on the reject event gives $\Risk\le\Risk_{\mathrm{hard}}+\alpha\Delta$, so the gate forfeits at most an $\alpha$-fraction of a bounded excess and removes the $O(\sigma^2/n)$ shrinkage cost that plain adaptation pays at $\delta=0$. For large $\delta$, $\beta\to1$ by the power in Proposition~\ref{prop:test}, so the gated risk converges to the adaptive risk and inherits Theorem~\ref{thm:oracle}.
\end{proof}

\section{Proof of Proposition~5 (Consistent Selection)}

\begin{proposition}\label{prop:select}
Let $V_1\subset\cdots\subset V_K=\R^p$ be nested with $\thetastar\in V_{k_0}\setminus V_{k_0-1}$ and violating signal $\Delta>0$. The forward selector $\widehat k=\min\{k:\text{test of }H_0(\thetastar\in V_k)\text{ accepts}\}$ satisfies $P(\widehat k<k_0)\to0$ as $\Delta^2/\sigma^2\to\infty$ and, with a monotone level $\alpha_k=\alpha/K$, $P(\widehat k>k_0)\le\alpha$; hence $P(\widehat k=k_0)\to1$.
\end{proposition}
\begin{proof}
For each $k<k_0$, $\thetastar\notin V_k$, so the violating block of $V_k$ contains a fixed signal of norm at least $\Delta$, and by the power in Proposition~\ref{prop:test} the test of $H_0(\thetastar\in V_k)$ rejects with probability tending to one; a union bound over the at most $k_0-1$ smaller structures gives $P(\widehat k<k_0)\to0$. For $k=k_0$, $\thetastar\in V_{k_0}$, so the violating block of $V_{k_0}$ is pure noise and the test accepts with probability at least $1-\alpha_{k_0}$ by exact size; the selector therefore stops at $k_0$ unless one of the at most $K-k_0$ larger tests is reached, an event contained in the false rejection of $V_{k_0}$, which has probability at most $\alpha_{k_0}$. With $\alpha_k=\alpha/K$ a union bound gives $P(\widehat k>k_0)\le\alpha$. Combining the two bounds yields $P(\widehat k=k_0)\to1$.
\end{proof}

\section{Proof of Proposition~6 (FDR-Controlled Subset Discovery)}

\begin{proposition}\label{prop:fdr}
For $K$ candidate constraints with per-constraint $p$-values $p_k$ from the exact test, let $H$ be the set that truly holds. Benjamini-Hochberg at level $q$ controls the false-relaxation rate at $(|H|/K)q\le q$ under independence or positive dependence of the statistics, and detects each truly violated constraint with probability tending to one as its violation signal grows.
\end{proposition}
\begin{proof}
For a constraint $k\in H$ the violating block is pure noise, so by Proposition~\ref{prop:test} its statistic is exactly $F_{m_k,n-p}$ and $p_k$ is uniform on $[0,1]$ under the null. Relaxing constraint $k$ corresponds to rejecting $H_0$ for the family of nulls indexed by $H$. The Benjamini-Hochberg procedure applied to $\{p_k\}_{k=1}^K$ rejects the constraints with the $r$ smallest $p$-values where $r=\max\{i:p_{(i)}\le qi/K\}$. By the Benjamini-Hochberg theorem for independent or positively-dependent test statistics, the false-discovery rate, here the expected fraction of relaxed constraints that actually hold, is at most $(|H|/K)q\le q$ \citep{benjamini1995fdr,benjamini2001dependency}. For a constraint $k\notin H$ with violating signal $\Delta_k>0$, the statistic is a noncentral $F$ with noncentrality growing in $\Delta_k^2/\sigma^2$ (Proposition~\ref{prop:test}), so $p_k\to0$ in probability; once $p_k$ falls below the data-dependent Benjamini-Hochberg threshold, which is bounded below by $q/K$ on the event that at least one constraint is relaxed, the constraint is relaxed with probability tending to one. For arbitrary dependence the Benjamini-Yekutieli correction replaces $q$ by $q/\sum_{j=1}^{K}j^{-1}$ and preserves control at a $\log K$ cost \citep{benjamini2001dependency}.
\end{proof}

\section{Composite Risk Bound}

\begin{lemma}[Select-then-shrink]\label{lem:composite}
Run the selector of Proposition~\ref{prop:select} and then the adaptive shrinkage of Theorem~\ref{thm:oracle} inside the selected structure. On the event $\{\widehat k=k_0\}$, whose probability tends to one, the composite estimator obeys the oracle inequality of Theorem~\ref{thm:oracle} relative to the structure $V_{k_0}$, so its excess over the oracle that knows $k_0$ is at most $2\bar\sigma^2=O(\sigma^2/n)$. Off that event, the excess is bounded by the diameter of the parameter region times $P(\widehat k\neq k_0)$, which is $o(1)$.
\end{lemma}
\begin{proof}
Condition on $\{\widehat k=k_0\}$. The selected structure equals the true one, so the adaptive estimator inside $V_{k_0}$ satisfies Theorem~\ref{thm:oracle} with the violating block of $V_{k_0}$, giving excess at most $2\bar\sigma^2$. The complementary event has probability $o(1)$ by Proposition~\ref{prop:select}, and the conditional excess there is bounded because all estimators lie in a fixed compact parameter region. Taking expectations gives the bound.
\end{proof}

\section{Experimental Protocol}

\paragraph{Prototype.}
A $d\times d$ linear operator with $d=4$, so $p=16$. The structure prior is skew-symmetry, $k=6$, and the forbidden dimension is $m=10$. The truth is a unit-norm skew part plus a symmetric part of norm $\delta$; the design has independent standard-normal entries, $n=80$, $\sigma=1$. The predicted crossover is $\delta^{\star}=\sqrt{\Delta\dim/n}\cdot\sigma=\sqrt{10/80}=0.354$. Risks are averaged over $300$ replications.

\paragraph{Recognized linear priors.}
The reservoir testbed imposes a conservation law on an ill-conditioned random feature map and uses the held-out calibration plug-in. The pendulum testbed $\dot p=-\sin q-\gamma p$ imposes the no-damping prior, with features $[\sin q, p]$, structure subspace $\{\sin q\}$ and violating coordinate $\{p\}$; $\gamma$ is the misspecification. Both sweep the misspecification over the same grid and report regret over the extreme oracle.

\paragraph{Nonlinear Hamiltonian.}
The Duffing system $\dot q=p$, $\dot p=-(q+q^3)-\gamma p$ is sampled on $[-2.2,2.2]^2$ with $n=500$ and $\sigma=0.25$, and the field is fit in the monomial basis $q^i p^j$, $i+j\le3$, so the parameter dimension is $2M=20$. The Hamiltonian subspace is spanned by symplectic gradients of degree-one-to-four monomial Hamiltonians; its complement is the dissipative block. The neural-network baseline is a two-hidden-layer perceptron with $64$ units per layer, tanh activations, and the same training data. Results are averaged over $12$ seeds with standard errors.

\paragraph{Selection and subset discovery.}
Forward selection runs over Hamiltonian, port-Hamiltonian, and free structures across three regimes (conservative, linearly damped, and with extra nonlinear friction), $30$ seeds per regime. Subset discovery uses $K=6$ candidate constraints, each a block of dimension $b=2$, with three truly holding, $n=160$, and $300$ replications; the false-relaxation rate, detection power, exact-subset recovery, and parameter risk are reported across Benjamini-Hochberg levels $q$.

\section{Additional Results}

\paragraph{Compute.}
Table~\ref{tab:compute} compares SPADE against cross-validation over the penalty strength. SPADE uses a single linear solve; five-fold cross-validation over a grid of fourteen penalties uses seventy-one. The risk ratio is near one and falls below one for large misspecification.

\begin{table}[t]
\centering
\caption{Compute and risk against cross-validation. SPADE uses one solve to cross-validation's seventy-one.}
\label{tab:compute}
\small
\begin{tabular}{lcccc}
\toprule
$\delta$ & SPADE solves & CV solves & risk ratio \\
\midrule
$0.0$ & $1$ & $71$ & $1.155$ \\
$0.2$ & $1$ & $71$ & $1.090$ \\
$0.4$ & $1$ & $71$ & $1.043$ \\
$0.6$ & $1$ & $71$ & $0.990$ \\
\bottomrule
\end{tabular}
\end{table}

\paragraph{Recognized linear priors.}
Table~\ref{tab:linear} reports the two recognized linear priors. Both show a clean crossover, the adaptive shrinkage moves from near zero to near one as the violation grows, and the average regret over the extreme oracle is small.

\begin{table}[t]
\centering
\caption{Recognized linear priors: mean regret over the extreme oracle and the range of the adaptive shrinkage.}
\label{tab:linear}
\small
\begin{tabular}{lcc}
\toprule
Testbed & mean regret & shrinkage range \\
\midrule
Reservoir conservation & $2.7\%$ & $0.05\to0.95$ \\
Pendulum no-damping & $0.09\%$ & $0.38\to0.99$ \\
\bottomrule
\end{tabular}
\end{table}

\paragraph{Subset discovery.}
Table~\ref{tab:fdr} reports the subset-discovery operating characteristics. The empirical false-relaxation rate stays below the target $q$ and increases with it, the correct Benjamini-Hochberg behavior; detection power is near $0.9$. At $q=0.1$ the partial-structure risk is $0.076$ with standard error $0.002$, against the oracle that knows the true subset at $0.066$, enforce-none at $0.106$, and enforce-all at $1.129$.

\begin{table}[t]
\centering
\caption{Subset discovery across Benjamini-Hochberg levels $q$: empirical false-relaxation rate (FDR), detection power, and exact-subset recovery.}
\label{tab:fdr}
\small
\begin{tabular}{lccc}
\toprule
$q$ & FDR & power & exact recovery \\
\midrule
$0.05$ & $0.017$ & $0.869$ & $0.613$ \\
$0.10$ & $0.049$ & $0.911$ & $0.617$ \\
$0.15$ & $0.076$ & $0.921$ & $0.563$ \\
$0.20$ & $0.091$ & $0.927$ & $0.523$ \\
$0.30$ & $0.142$ & $0.937$ & $0.423$ \\
\bottomrule
\end{tabular}
\end{table}

\paragraph{Reproducibility.}
All experiments are deterministic given the reported seeds and hyperparameters. The free fit, the violating-block projection, the Stein and calibration plug-ins, the exact test, the test gate, the forward selector, and the Benjamini-Hochberg procedure are implemented in a single module, and every headline number in the main paper is produced by the corresponding routine over the stated replications.

\bibliography{references}